\documentclass[conference]{IEEEtran}
\IEEEoverridecommandlockouts
\usepackage{cite}
\usepackage{amsmath,amssymb,amsfonts}
\usepackage{algorithmic}
\usepackage{booktabs}
\usepackage{graphicx}
\usepackage{textcomp}
\usepackage{xcolor}
\def\BibTeX{{\rm B\kern-.05em{\sc i\kern-.025em b}\kern-.08em
    T\kern-.1667em\lower.7ex\hbox{E}\kern-.125emX}}
\begin{document}

\title{LRD-Net: A Lightweight Real-Centered Detection Network for Cross-Domain Face Forgery Detection}

\author{\IEEEauthorblockN{Xuecen Zhang}
\IEEEauthorblockA{\textit{Department of Computer and Data Sciences} \\
\textit{Case Western Reserve University}\\
Cleveland, OH, USA \\
xxz1037@case.edu}
\and
\IEEEauthorblockN{Vipin Chaudhary}
\IEEEauthorblockA{\textit{Department of Computer and Data Sciences} \\
\textit{Case Western Reserve University}\\
Cleveland, OH, USA \\
vxc204@case.edu}
}

\maketitle
\IEEEpubid{\makebox[\columnwidth]{979-8-3315-7310-2/26/\$31.00~©2026 IEEE \hfill}%
\hspace{\columnsep}\makebox[\columnwidth]{}}
\IEEEpubidadjcol
\begin{abstract}
The rapid advancement of diffusion-based generative models has made face forgery detection a critical challenge in digital forensics. Current detection methods face two fundamental limitations: poor cross-domain generalization when encountering unseen forgery types, and substantial computational overhead that hinders deployment on resource-constrained devices. We propose LRD-Net (Lightweight Real-centered Detection Network), a novel framework that addresses both challenges simultaneously. Unlike existing dual-branch approaches that process spatial and frequency information independently, LRD-Net adopts a sequential frequency-guided architecture where a lightweight Multi-Scale Wavelet Guidance Module generates attention signals that condition a MobileNetV3-based spatial backbone. This design enables effective exploitation of frequency-domain cues while avoiding the redundancy of parallel feature extraction. Furthermore, LRD-Net employs a real-centered learning strategy with exponential moving average prototype updates and drift regularization, anchoring representations around authentic facial images rather than modeling diverse forgery patterns. Extensive experiments on the DiFF benchmark demonstrate that LRD-Net achieves state-of-the-art cross-domain detection accuracy, consistently outperforming existing methods. Critically, LRD-Net accomplishes this with only 2.63M parameters—approximately 9× fewer than conventional approaches—while achieving over 8× faster training and nearly 10× faster inference. These results demonstrate that robust cross-domain face forgery detection can be achieved without sacrificing computational efficiency, making LRD-Net suitable for real-time deployment in mobile authentication systems and resource-constrained environments.
\end{abstract}


\section{Introduction}



The rapid advancement of generative artificial intelligence, particularly diffusion-based models, has made image forgery detection a critical research area in computer vision and digital forensics. With the emergence of powerful generation tools such as Stable Diffusion~\cite{rombach2022high}, DALL-E~\cite{ramesh2022hierarchical}, and Midjourney, synthesizing highly realistic facial images has become remarkably easy and accessible to the general public. While these technologies offer creative benefits, they simultaneously pose severe threats to digital security and public trust. Malicious actors can exploit forged facial images for identity theft, enabling unauthorized access to personal accounts and financial systems. In social media and journalism, fabricated images of public figures can fuel disinformation campaigns and political manipulation, undermining democratic processes. Furthermore, synthetic faces can be weaponized for social engineering attacks, where convincing fake identities deceive individuals or organizations into revealing sensitive information. These escalating risks underscore the urgent need for robust and practical face forgery detection methods.

Current research in face forgery detection faces two fundamental challenges that limit real-world deployment. The first challenge concerns \textbf{cross-domain generalization}. Modern diffusion-based face forgeries exhibit diverse manipulation strategies, among which \textit{Face Editing (FE)}, \textit{Image-to-Image translation (I2I)}, and \textit{Text-to-Image generation (T2I)} are widely regarded as representative categories, corresponding to attribute-level modification of real faces, transformation of source facial images, and synthesis of entirely new faces from textual descriptions, respectively. Although these forgery types share a common goal of producing realistic faces, they exhibit substantially different visual characteristics and generation artifacts.
Existing detection methods achieve near-perfect accuracy when training and testing are conducted on the same forgery type—referred to as \textit{in-domain} evaluation. However, performance degrades significantly when the detector is evaluated on unseen forgery types, a setting commonly referred to as \textit{cross-domain} evaluation~\cite{haliassos2021lips}. This limitation is particularly problematic given the rapid evolution of generative models: new diffusion architectures and fine-tuned variants emerge frequently, rendering detectors trained on older forgery types ineffective against novel manipulations. A detector that fails to generalize across forgery domains provides a false sense of security and cannot serve as a reliable defense in practice.

The second challenge concerns computational efficiency. Several prior works improve cross-domain robustness by adopting deeper backbones, multi-branch feature extraction, or complex feature fusion strategies~\cite{luo2021generalizing, qian2020thinking, zhang2025dual}. For example, RCDN achieves strong cross-domain performance by anchoring representations to authentic images and jointly modeling spatial and frequency information via a dual-branch architecture. However, such designs typically rely on heavyweight backbones (e.g., Xception) and parallel pipelines, leading to high parameter counts and substantial computational overhead.
These costs hinder practical deployment, especially in resource-constrained scenarios such as mobile authentication systems, where strict limits on computation, memory, and energy apply. Consequently, there is a clear need for lightweight detectors that maintain strong cross-domain robustness while significantly reducing computational complexity, enabling real-world and edge-device deployment.

To address these challenges, we propose \textbf{LRD-Net (Lightweight Real-centered Detection Network)}, a novel framework that simultaneously improves cross-domain generalization and computational efficiency. Unlike existing dual-branch approaches that process spatial and frequency information independently, LRD-Net adopts a \textbf{sequential frequency-guided design} in which lightweight frequency analysis generates guidance signals that steer spatial feature extraction. This design enables the model to exploit frequency-domain cues for localization and emphasis, while avoiding the redundancy and overhead of parallel feature fusion.
Furthermore, LRD-Net employs a \textbf{real-centered learning strategy} that anchors the representation space around authentic facial images rather than attempting to model the diverse and rapidly evolving patterns of forged content. By emphasizing the intrinsic consistency of real faces and enforcing stability in the learned representation, LRD-Net achieves robust generalization to unseen forgery types. Importantly, the overall architecture is deliberately designed to be lightweight, making it suitable for deployment in resource-constrained environments such as mobile and embedded systems.

In summary, the contributions of this work are threefold:
\begin{itemize}
    \item We propose \textbf{LRD-Net}, a lightweight and real-centered face forgery detection framework that adopts a sequential frequency-guided architecture to improve cross-domain generalization across diffusion-based manipulations.
    
    \item We design LRD-Net to be computationally efficient, reducing model parameters by approximately 9$\times$ compared to existing parallel dual-branch methods while maintaining competitive detection performance.
    
    \item We conduct extensive experiments on the \textbf{DiFF benchmark}~\cite{cheng2024diffusion}. Experimental results demonstrate that LRD-Net achieves state-of-the-art cross-domain robustness with significantly lower computational cost.
\end{itemize}

\section{Related Work}
Face forgery detection has received increasing attention with the rapid advancement of generative models. Early approaches relied on handcrafted features and traditional signal processing techniques to identify inconsistencies in noise statistics, compression artifacts, or local image patterns~\cite{farid2009image}. While effective for conventional image manipulations, these methods struggle against modern generative models that produce highly realistic and visually coherent facial images~\cite{mirsky2021creation}.
With the rise of deep learning, convolutional neural networks (CNNs) have become the dominant paradigm for face forgery detection. Large-scale benchmarks such as DiFF~\cite{cheng2024diffusion} have facilitated the development of CNN-based detectors using standard backbones including Xception~\cite{chollet2017xception}, ResNet variants, and EfficientNet~\cite{tan2019efficientnet}. These models typically achieve near-perfect accuracy under in-domain evaluation, where training and testing samples originate from the same forgery distribution. However, two fundamental challenges continue to limit their practical deployment.

The first challenge is cross-domain generalization. CNN-based detectors often overfit to forgery-specific artifacts in the training data, leading to significant performance degradation on unseen forgery types~\cite{haliassos2021lips}. This problem is exacerbated in diffusion-based generation, where diverse pipelines and rapidly evolving models introduce heterogeneous artifacts, causing detectors trained on fixed manipulations to generalize poorly in real-world settings.
To address this issue, prior work has explored frequency-domain analysis, leveraging systematic spectral artifacts introduced by generative models~\cite{dzanic2020fourier}. Frequency-aware and multi-branch methods that jointly model spatial and spectral cues have shown improved robustness~\cite{qian2020thinking, luo2021generalizing}. More recent approaches incorporate diffusion-specific characteristics, such as reconstruction-based detection~\cite{wang2023dire} and contrastive learning for universal diffusion detection~\cite{chen2024drct}, further enhancing generalization to unseen generation models.

Despite these advances, existing approaches introduce a second major challenge: model complexity and computational overhead. Many frequency-aware and diffusion-specific methods—including RCDN—rely on deep backbone networks, reconstruction pipelines, or parallel feature extraction strategies, resulting in tens of millions of parameters and substantial computational cost~\cite{qian2020thinking, wang2023dire}. While effective for improving cross-domain robustness, such designs are difficult to deploy in resource-constrained environments such as mobile devices and embedded systems, where memory footprint, latency, and energy consumption are critical concerns.

In summary, prior work in face forgery detection reveals a clear trade-off between cross-domain robustness and computational efficiency. While frequency-aware and real-centered frameworks demonstrate that leveraging stable properties of authentic images is a promising direction for generalization, their reliance on heavyweight architectures limits practical applicability. These limitations motivate the development of a lightweight yet robust detection framework, which we address with the proposed LRD-Net.

\section{Proposed Method}

\begin{figure*}[t]
    \centering
    \includegraphics[width=0.9\linewidth]{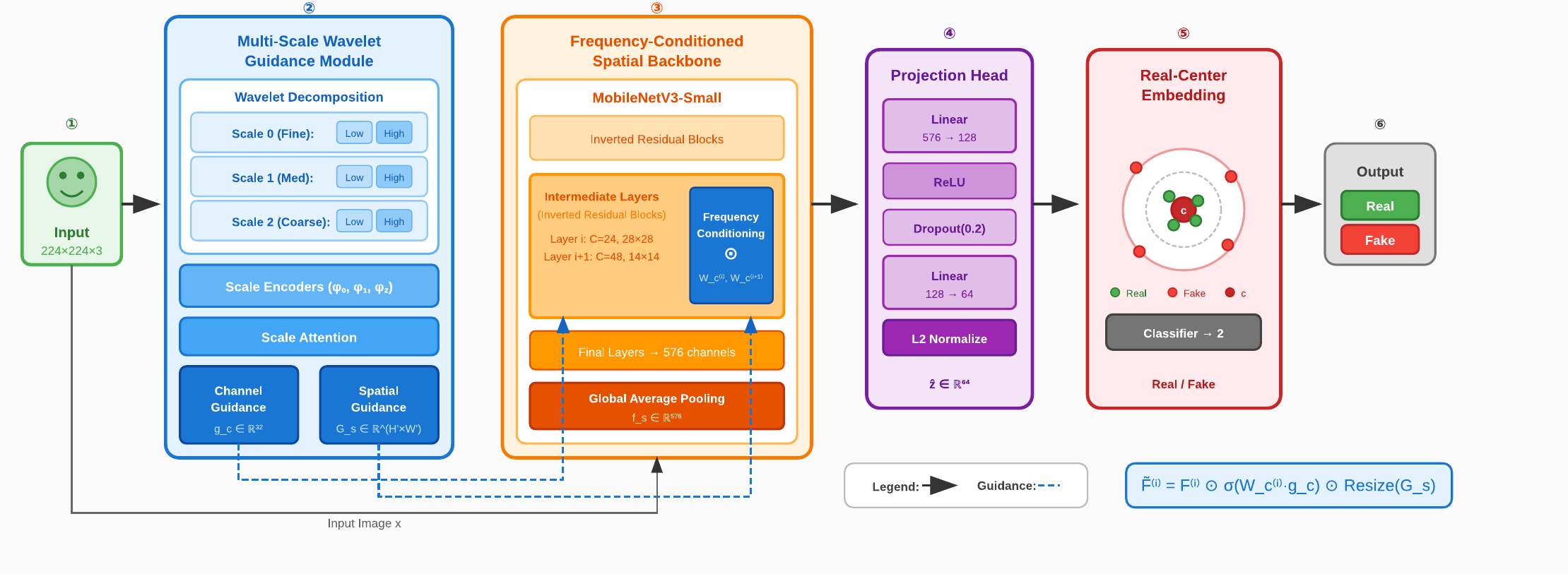}
    \caption{Overall pipeline of the proposed method.}
    \label{fig:architecture}
\end{figure*}

\subsection{Overview}
The overall architecture of LRD-Net is illustrated in Fig.~\ref{fig:architecture}. Unlike conventional dual-branch methods that process spatial and frequency information in parallel, LRD-Net adopts a \textbf{sequential frequency-guided paradigm}.


Given an input face image $\mathbf{x} \in \mathbb{R}^{H \times W \times 3}$, LRD-Net processes it through four stages: (1) a lightweight Multi-Scale Wavelet Guidance Module generates frequency-derived guidance signals; (2) these signals condition a MobileNetV3-based spatial backbone to emphasize forgery-relevant regions and channels; (3) extracted features are projected into a compact embedding space anchored around a real-center prototype; and (4) a linear classifier produces the final prediction. The training objective combines classification loss with real-centered constraints and prototype drift regularization.

\subsection{Model Architecture}

\subsubsection{Multi-Scale Wavelet Guidance Module}
A fundamental observation in face forgery detection is that different generation methods introduce artifacts at different frequency scales~\cite{zhang2019detecting, schwarz2021frequency}. Consequently, a fixed single-scale frequency analysis lacks the flexibility to capture such diverse artifacts. To address this limitation, we propose the \textbf{Multi-Scale Wavelet Guidance Module (MSWGM)}, which decomposes the input image into multiple frequency bands and learns to emphasize the most discriminative scales for forgery detection.

Given an input image $\mathbf{x}$, MSWGM performs a wavelet-like multi-scale decomposition at $L$ scales (default $L=3$). Rather than applying a full discrete wavelet transform, we adopt an efficient approximation using Gaussian filtering with scale-dependent kernel sizes. Specifically, at each scale $l \in \{0, 1, \ldots, L-1\}$, the low-frequency component $\mathbf{x}_l^{\mathrm{low}}$ is obtained via Gaussian smoothing, and the corresponding high-frequency detail is computed as:
\begin{equation}
\mathbf{x}_l^{\mathrm{high}} = \mathbf{x} - \mathbf{x}_l^{\mathrm{low}}.
\end{equation}

The low- and high-frequency components are concatenated and passed through a lightweight encoder $\phi_l(\cdot)$ to extract scale-specific representations:
\begin{equation}
\mathbf{f}_l = \phi_l([\mathbf{x}_l^{\mathrm{low}} \, \| \, \mathbf{x}_l^{\mathrm{high}}]), 
\quad \mathbf{f}_l \in \mathbb{R}^{d_f}.
\end{equation}

To adaptively determine the importance of each frequency scale, we introduce a \textbf{scale attention mechanism} that assigns a normalized weight to each scale based on its discriminative contribution:
\begin{equation}
[\alpha_0, \alpha_1, \ldots, \alpha_{L-1}] 
= \mathrm{Softmax}\left(
g([\mathbf{f}_0 \, \| \, \mathbf{f}_1 \, \| \, \ldots \, \| \, \mathbf{f}_{L-1}])
\right),
\end{equation}
where $g(\cdot)$ denotes a lightweight attention network. This mechanism allows the model to dynamically focus on frequency bands that are most informative for a given input.

Based on the attended multi-scale representations, MSWGM outputs two types of guidance signals:
\begin{itemize}
    \item \textbf{Channel Guidance} $\mathbf{g}_c \in \mathbb{R}^{d_g}$, which indicates which feature channels should be emphasized during spatial feature extraction;
    \item \textbf{Spatial Guidance} $\mathbf{G}_s \in \mathbb{R}^{H' \times W'}$, which highlights spatial regions that are likely to contain forgery artifacts.
\end{itemize}

\subsubsection{Frequency-Conditioned Spatial Backbone}
The spatial backbone is responsible for extracting semantic facial features from the input image. Instead of employing a heavy architecture such as Xception~\cite{chollet2017xception} (approximately 22M parameters), we adopt MobileNetV3-Small~\cite{howard2019searching} (approximately 2.5M parameters) as the base network. MobileNetV3-Small is specifically designed for mobile and edge deployment, making it well suited for practical forgery detection scenarios with limited computational resources.

The key innovation lies in conditioning the spatial backbone with frequency guidance signals, establishing a \textbf{sequential information flow} from frequency analysis to spatial feature extraction. Rather than treating frequency and spatial information as independent modalities, the proposed design uses frequency-derived cues to dynamically modulate spatial feature learning. Conditioning is applied at intermediate layers of the backbone, where mid-level semantic representations emerge and are most informative for forgery detection.

Let $\mathbf{F}^{(i)} \in \mathbb{R}^{C_i \times H_i \times W_i}$ denote the feature map produced by the $i$-th group of inverted residual blocks in MobileNetV3-Small. Frequency conditioning is performed by jointly applying channel-wise and spatial guidance:
\begin{equation}
    \tilde{\mathbf{F}}^{(i)} =
    \mathbf{F}^{(i)} \odot
    \sigma\!\left(\mathbf{W}_c^{(i)} \mathbf{g}_c\right)
    \odot
    \mathrm{Resize}\!\left(\mathbf{G}_s\right),
\end{equation}
where $\mathbf{W}_c^{(i)} \in \mathbb{R}^{C_i \times d_g}$ is a learnable projection matrix that adapts the channel guidance vector $\mathbf{g}_c$ to match the feature dimension at stage $i$, $\sigma(\cdot)$ denotes the sigmoid activation, $\mathrm{Resize}(\cdot)$ spatially interpolates the guidance map $\mathbf{G}_s$ to the resolution $(H_i, W_i)$, and $\odot$ represents element-wise multiplication.

We apply this conditioning at two intermediate groups of inverted residual blocks in MobileNetV3-Small, corresponding to feature resolutions of $28\times28$ and $14\times14$ with channel dimensions of 24 and 48, respectively. These stages capture mid-level semantic features that balance spatial detail and semantic abstraction, making them particularly suitable for frequency-guided modulation. The final feature map is then processed by global average pooling to produce a compact spatial representation $\mathbf{f}_s \in \mathbb{R}^{576}$.

By conditioning the spatial backbone in this manner, LRD-Net effectively integrates frequency-domain cues into semantic feature extraction while maintaining a lightweight architecture, enabling robust and efficient face forgery detection.

\subsubsection{Projection Head and Real-Center Embedding}

The spatial features are projected into a compact embedding space via a projection head:
\begin{equation}
\mathbf{z} = \psi(\mathbf{f}_s), \quad \mathbf{z} \in \mathbb{R}^{d_e},
\end{equation}
where $d_e = 64$ denotes the embedding dimension. The resulting embedding is then $\ell_2$-normalized.

We maintain a prototype vector $\mathbf{c}$ representing the center of authentic face embeddings. Unlike prior approaches such as RCDN, which treat the center as a fully learnable parameter updated via gradient descent, LRD-Net updates the real center using an \textbf{exponential moving average (EMA)} of real sample embeddings:
\begin{equation}
\mathbf{c}^{(t)} = \mu \cdot \mathbf{c}^{(t-1)} + (1 - \mu) \cdot \bar{\mathbf{z}}^{(t)}_{\mathrm{real}},
\end{equation}
where $\bar{\mathbf{z}}^{(t)}_{\mathrm{real}}$ denotes the mean of the $\ell_2$-normalized embeddings of real samples in the current mini-batch, and $\mu = 0.99$ is the momentum coefficient. After each update, $\mathbf{c}$ is $\ell_2$-normalized to remain in the same embedding space.

This momentum-based update yields a more stable estimate of the real embedding center. In contrast to a fully learnable center, which may drift arbitrarily under gradient updates and overfit to training-domain forgery patterns, the EMA-based center evolves smoothly and captures the intrinsic structure of real face embeddings. As a result, the learned representation exhibits improved robustness and generalization to unseen forgery domains.

\subsection{Loss Function Design}
\label{sec:loss}

The training objective of LRD-Net consists of three complementary loss terms:
\begin{equation}
    \mathcal{L} = \mathcal{L}_{\mathrm{cls}} + \lambda_c \mathcal{L}_{\mathrm{center}} + \lambda_d \mathcal{L}_{\mathrm{drift}},
\end{equation}
where $\lambda_c$ and $\lambda_d$ control the relative contributions of the real-centered constraint and prototype regularization.

\subsubsection{Classification Loss}
To ensure discriminative learning within the training distribution, we employ the standard binary cross-entropy loss:
\begin{equation}
    \mathcal{L}_{\mathrm{cls}} = -\frac{1}{N} \sum_{i=1}^{N} \left[
    y_i \log p_i + (1 - y_i) \log (1 - p_i)
    \right],
\end{equation}
where $N$ is the batch size, $y_i \in \{0,1\}$ denotes the ground-truth label (real or fake), and $p_i$ is the predicted probability of being fake.

\subsubsection{Real-Centered Loss}
To enforce a structured embedding space, we introduce a real-centered loss that pulls authentic samples toward the center while pushing forged samples away:
\begin{equation}
    \mathcal{L}_{\mathrm{center}} =
    \frac{1}{|R|} \sum_{\hat{\mathbf{z}}_i \in R} d(\hat{\mathbf{z}}_i, \mathbf{c})^2
    +
    \frac{1}{|F|} \sum_{\hat{\mathbf{z}}_j \in F}
    \max\!\left(0, m - d(\hat{\mathbf{z}}_j, \mathbf{c})\right)^2,
\end{equation}
where $R$ and $F$ denote the sets of real and fake embeddings in the mini-batch, $d(\cdot,\cdot)$ is the Euclidean distance, and $m$ is a margin hyperparameter. This formulation encourages real embeddings to form a compact cluster around the prototype, while enforcing a minimum separation between fake samples and the real center without constraining their internal distribution.

\subsubsection{Prototype Drift Regularization}
Unlike prior real-centered approaches such as RCDN, which introduce an explicit separation loss based on relative sample distances, LRD-Net directly regularizes the stability of the real prototype. Specifically, we introduce a \textbf{prototype drift regularization} that penalizes excessive changes in the center position:
\begin{equation}
    \mathcal{L}_{\mathrm{drift}} =
    \max\!\left(0, \|\mathbf{c}^{(t)} - \mathbf{c}^{(t-1)}\|_2 - \delta \right),
\end{equation}
where $\mathbf{c}^{(t)}$ and $\mathbf{c}^{(t-1)}$ denote the prototype at consecutive update steps, and $\delta$ is a drift threshold.

This loss explicitly encourages a stable real prototype during training. Large prototype drift indicates that the model is adapting to spurious or forgery-specific patterns, whereas a stable center reflects the learning of invariant characteristics of authentic faces. When combined with the margin-based real-centered loss, this regularization implicitly enforces separation between real and fake samples, rendering explicit separation losses unnecessary. As a result, LRD-Net achieves more stable training dynamics and improved cross-domain generalization with fewer hyperparameters.

\subsection{Lightweight Analysis}
To demonstrate the lightweight design of LRD-Net, we compare it with RCDN, which achieves state-of-the-art performance on the DiFF benchmark.

As shown in Table~\ref{tab:param_count}, RCDN contains 24.48M parameters due to its Xception backbone and parallel frequency feature extraction branch, whereas LRD-Net contains only 2.63M parameters, achieving a $9.3\times$ reduction.
This reduction results from three design choices: replacing Xception with MobileNetV3-Small, using a guidance-only frequency module instead of full feature extraction, and adopting a compact projection head. Together, these choices significantly reduce model size without compromising detection performance.

\begin{table}[t]
\centering
\caption{Parameter count comparison between RCDN and LRD-Net.}
\label{tab:param_count}
\begin{tabular}{lcc}
\toprule
\textbf{Component} & \textbf{RCDN} & \textbf{LRD-Net} \\
\midrule
Spatial Backbone & 22,855,952 & 2,539,376 \\
Frequency Branch & 378,944  & 7,780  \\
Projection Head  & 1,245,824 (2304$\rightarrow$128) & 82,112 (576$\rightarrow$64) \\
Classifier       & 258 & 130 \\
\midrule
\textbf{Total}    & \textbf{24,481,106} & \textbf{2,629,398} \\
\midrule
\textbf{Reduction} & baseline & \textbf{9.3$\times$} \\
\bottomrule
\end{tabular}
\end{table}

Beyond parameter count, memory footprint is critical for practical deployment. Table~\ref{tab:memory_analysis} compares the storage requirements of RCDN and LRD-Net under different numerical precisions.
Under FP32 precision, RCDN requires 97.9~MB of storage, whereas LRD-Net requires only 10.5~MB. This advantage remains under reduced precision: LRD-Net occupies 5.3~MB with FP16 and 2.6~MB with INT8, compared to 49.0~MB and 24.5~MB for RCDN. The consistent $9.3\times$ reduction indicates that LRD-Net’s efficiency primarily stems from architectural design rather than quantization.
With INT8 quantization, LRD-Net meets the memory constraints of mobile and embedded systems, whereas RCDN remains impractical for such deployments.

\begin{table}[t]
\centering
\caption{Memory footprint comparison between RCDN and LRD-Net under different numerical precisions.}
\label{tab:memory_analysis}
\begin{tabular}{lccc}
\toprule
\textbf{Metric} & \textbf{RCDN} & \textbf{LRD-Net} & \textbf{Calculation} \\
\midrule
Parameters & 24,481,106 & 2,629,398 & -- \\
FP32 Size  & 97.9 MB & 10.5 MB & Params $\times$ 4 bytes \\
FP16 Size  & 49.0 MB & 5.3 MB & Params $\times$ 2 bytes \\
INT8 Size  & 24.5 MB & 2.6 MB & Params $\times$ 1 byte \\

\bottomrule
\end{tabular}
\end{table}

\section{Experiments}

\subsection{Dataset}
We conduct experiments on the DiFF dataset~\cite{cheng2024diffusion}, a large-scale benchmark for diffusion-generated facial forgery detection that covers diverse generation methods and manipulation categories.
Following RCDN~\cite{xinzhang}, we adopt their carefully curated subset of DiFF, which focuses on three representative forgery categories: Face Edit (FE), Image-to-Image (I2I), and Text-to-Image (T2I). This subset excludes low-quality samples such as images that fail face detection, exhibit obvious visual artifacts (e.g., distorted eyes, asymmetric facial features), or contain severely blurred facial regions. Each category contains 10,000 images for training and 2,000 images for testing, with a balanced distribution of real and forged samples. For cross-domain evaluation, models are trained on one category and tested on all categories to assess generalization capability.

\subsection{In-domain Performance}
Table.~\ref{tab:indomain} presents the in-domain detection accuracy where models are trained and tested on the same subset. LRD-Net achieves competitive performance across all three generation categories, with accuracies of 0.9940 on FE, 0.9945 on I2I, and 0.9980 on T2I. These results are comparable to RCDN and substantially outperforms earlier approaches such as Xception, EfficientNet, and ResNet-34, and significantly surpass XcepKNN. The marginal difference between LRD-Net and RCDN in the in-domain setting (less than 0.6\% on average) demonstrates that our lightweight architecture retains strong discriminative capability when domain shift is absent.

\begin{table}[htbp]
\caption{In-domain detection accuracy on the DiFF dataset. Training and testing are conducted on the same subset.}
\centering
\begin{tabular}{|l|c|c|c|}
\hline
\textbf{Method} & \textbf{Train FE} & \textbf{Train I2I} & \textbf{Train T2I} \\
\hline
Xception~\cite{chollet2017xception}     & 0.9895 & 0.9860 & 0.9905 \\\hline
EfficientNet~\cite{koonce2021efficientnet} & 0.9980 & 0.9880 & 0.9930 \\\hline
ResNet-34~\cite{xie2017aggregated}    & 0.9890 & 0.9835 & 0.9900 \\\hline
XcepKNN~\cite{gilles2025xcepknn}       & 0.8132 & 0.7773 & 0.7803 \\\hline
DIRE~\cite{wang2023dire}          & 0.9820 & 0.9655 & 0.9850 \\\hline
RCDN~\cite{xinzhang} & 0.9995 & 0.9975 & 0.9990 \\ \hline
\textbf{LRD-Net} & 0.9940 &0.9945 & 0.9980\\
\hline
\end{tabular}
\label{tab:indomain}
\end{table}

\subsection{Cross-domain Performance}
Table.~\ref{tab:crossdomain_raw} reports cross-domain detection accuracy on the DIFF dataset, where each model is trained on one manipulation subset and evaluated on the remaining subsets. Overall, all methods exhibit a noticeable performance drop when transferred across domains, highlighting the intrinsic domain gap among different forgery generation processes. Our LRD-Net consistently outperforms all competing methods in cross-domain detection accuracy, demonstrating strong robustness to domain shifts across different forgery generation processes. When trained on I2I, LRD-Net achieves the highest cross-domain average accuracy of 0.9363, surpassing the strongest baseline RCDN by a clear margin. Notably, LRD-Net maintains high performance regardless of the training domain, achieving cross-domain averages of 0.9012 (FE) and 0.9052 (T2I), whereas other methods exhibit larger performance fluctuations across training settings.
These results confirm that LRD-Net is particularly well suited for cross-domain forgery detection, where robustness to distribution shifts is critical for real-world deployment.

\begin{table}[htbp]
\caption{Cross-domain detection accuracy on the DiFF dataset. Rows correspond to training subsets, and columns to testing subsets. In-domain diagonals are omitted; the rightmost column reports the cross-domain average.}
\centering
\begin{tabular}{|l|c|c|c|c|}
\hline
\textbf{Method (Train)} & \textbf{Test FE} & \textbf{Test I2I} & \textbf{Test T2I} & \textbf{Cross Avg} \\
\hline
Xception (FE)   & ---    & 0.8675 & 0.8850 & 0.8763 \\
Xception (I2I)  & 0.8024 & ---    & 0.9470 & 0.8747 \\
Xception (T2I)  & 0.7825 & 0.9395 & ---    & 0.8610 \\
\hline
EfficientNet (FE)   & ---    & 0.8660 & 0.9125 & 0.8892 \\
EfficientNet (I2I)  & 0.8405 & ---    & 0.9675 & 0.9040 \\
EfficientNet (T2I)  & 0.7395 & 0.9225 & ---    & 0.8310 \\
\hline
ResNet-34 (FE)   & ---    & 0.8195 & 0.8460 & 0.8327 \\
ResNet-34 (I2I)  & 0.8325 & ---    & 0.9625 & 0.8975 \\
ResNet-34 (T2I)  & 0.7430 & 0.9345 & ---    & 0.8387 \\
\hline
XcepKNN (FE)   & ---    & 0.6480 & 0.6164 & 0.6322 \\
XcepKNN (I2I)  & 0.6224 & ---    & 0.7410 & 0.6817 \\
XcepKNN (T2I)  & 0.5625 & 0.7638 & ---    & 0.6632 \\
\hline
DIRE (FE)   & ---    & 0.8695 & 0.8700 & 0.8698 \\
DIRE (I2I)  & 0.8560 & ---    & 0.9280 & 0.8920 \\
DIRE (T2I)  & 0.7985 & 0.9205 & ---    & 0.8595 \\
\hline
RCDN (FE)   & ---    & 0.8705 & 0.9125 & 0.8915 \\
RCDN (I2I)  & 0.8660 & ---    & 0.9810 & 0.9235 \\
RCDN (T2I)  & 0.8385 & 0.9560 & ---    & 0.8972 \\
\hline
LRD-Net (FE)   & ---   & 0.8805 & 0.9220 & 0.9012 \\
LRD-Net (I2I)  & 0.8755 & ---  & 0.9970 & 0.9363 \\
LRD-Net (T2I)  & 0.8355 & 0.9750 & ---  & 0.9052 \\
\hline
\end{tabular}
\label{tab:crossdomain_raw}
\end{table}

\subsection{Time Efficiency}
\begin{figure}[htbp]
    \centering
    \includegraphics[width=0.95\linewidth]{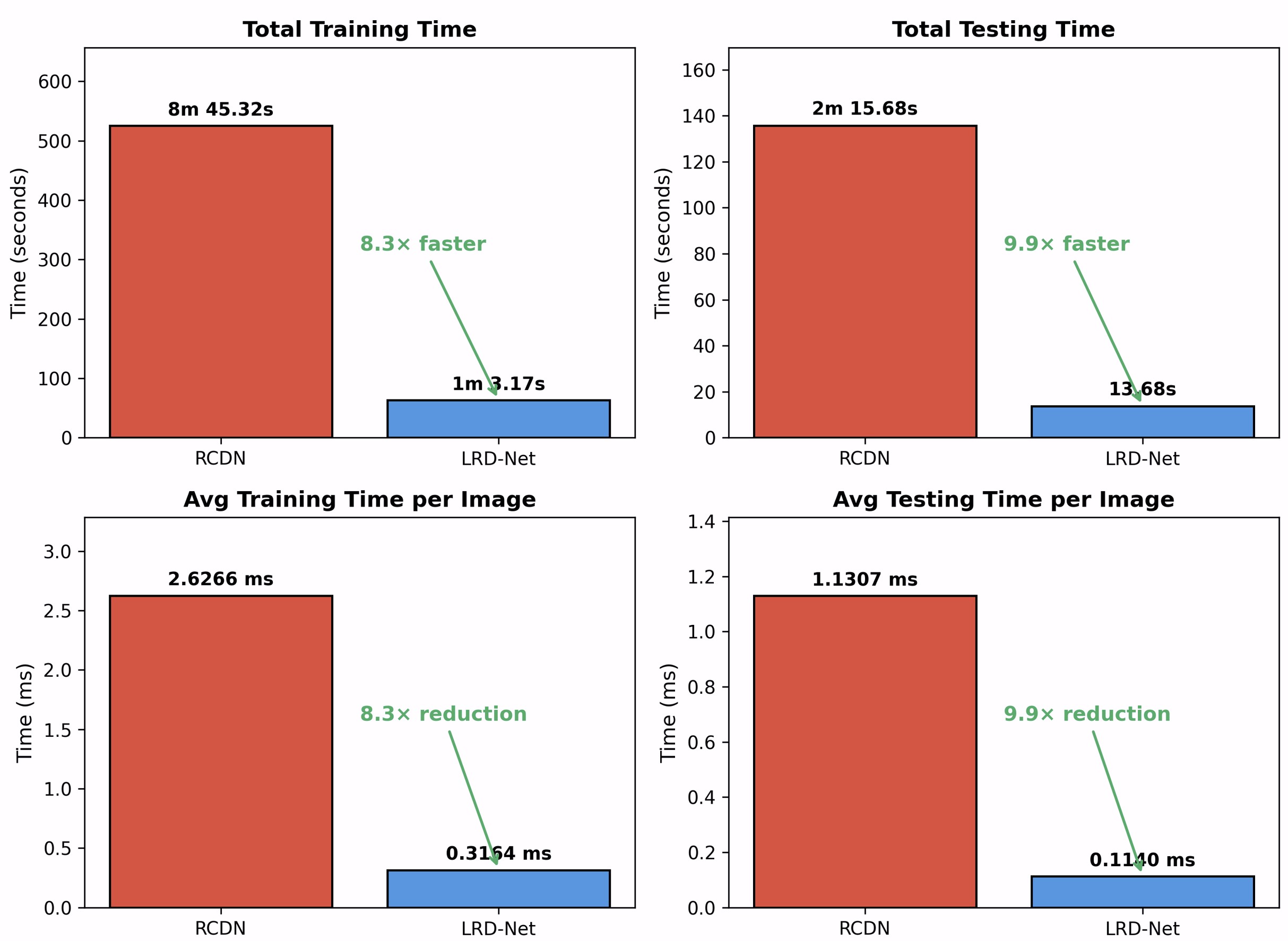}
    \caption{Time efficiency comparison: LRD-Net vs RCDN. All the experiments are conducted on RTX 4090. }
    \label{fig:te}
\end{figure}
Figure.~\ref{fig:te} presents a comprehensive comparison of computational efficiency between LRD-Net and RCDN. For total training time, LRD-Net completes training in 1 minute 3.17 seconds compared to RCDN's 8 minutes 45.32 seconds, achieving an 8.3× speedup. The testing efficiency improvement is even more pronounced, with LRD-Net requiring only 13.68 seconds versus RCDN's 2 minutes 15.68 seconds, representing a 9.9× acceleration. At the per-image level, LRD-Net processes each training image in 0.3164 ms compared to RCDN's 2.6266 ms, and each testing image in 0.1140 ms versus 1.1307 ms. The significant reduction in computational overhead makes LRD-Net particularly suitable for real-time deployment scenarios and resource-constrained environments, while maintaining competitive detection accuracy.

\section{Conclusion}
This paper presented LRD-Net, a lightweight real-centered detection network designed to address the dual challenges of cross-domain generalization and computational efficiency in face forgery detection. The proposed framework introduces three key innovations: a sequential frequency-guided architecture that replaces heavyweight dual-branch designs with lightweight guidance signals, a Multi-Scale Wavelet Guidance Module that adaptively emphasizes discriminative frequency bands, and an EMA-based real-centered learning strategy with prototype drift regularization that stabilizes training and improves generalization. Experimental results on the DiFF benchmark demonstrate that LRD-Net achieves superior cross-domain detection accuracy compared to existing methods. These accuracy gains are achieved while reducing model parameters by 9.3×, training time by 8.3×, and inference time by 9.9×. Our findings demonstrate that the trade-off between detection robustness and computational efficiency can be effectively mitigated through careful architectural design.

\bibliographystyle{IEEEtran}
\bibliography{ref}

@inproceedings{rombach2022high,
   title={High-resolution image synthesis with latent diffusion models},
   author={Rombach, Robin and Blattmann, Andreas and Lorenz, Dominik and Esser, Patrick and Ommer, Bj{\"o}rn},
   booktitle={Proceedings of the IEEE/CVF Conference on Computer Vision and Pattern Recognition},
   pages={10684--10695},
   year={2022}
 }

@article{ramesh2022hierarchical,
   title={Hierarchical text-conditional image generation with clip latents},
   author={Ramesh, Aditya and Dhariwal, Prafulla and Nichol, Alex and Chu, Casey and Chen, Mark},
   journal={arXiv preprint arXiv:2204.06125},
   year={2022}
 }

@inproceedings{cheng2024diffusion,
   title={Diffusion facial forgery detection},
   author={Cheng, Harry and Guo, Yangyang and Wang, Tianyi and Nie, Liqiang and Kankanhalli, Mohan},
   booktitle={Proceedings of the 32nd ACM International Conference on Multimedia},
   pages={5939--5948},
   year={2024}
 }

@inproceedings{haliassos2021lips,
   title={Lips don't lie: A generalisable and robust approach to face forgery detection},
   author={Haliassos, Alexandros and Vougioukas, Konstantinos and Petridis, Stavros and Pantic, Maja},
   booktitle={Proceedings of the IEEE/CVF conference on computer vision and pattern recognition},
   pages={5039--5049},
   year={2021}
 }

@inproceedings{luo2021generalizing,
   title={Generalizing face forgery detection with high-frequency features},
  author={Luo, Yuchen and Zhang, Yong and Yan, Junchi and Liu, Wei},
  booktitle={Proceedings of the IEEE/CVF conference on computer vision and pattern recognition},
  pages={16317--16326},
  year={2021}
 }

@inproceedings{qian2020thinking,
   title={Thinking in frequency: Face forgery detection by mining frequency-aware clues},
   author={Qian, Yuyang and Yin, Guojun and Sheng, Lu and Chen, Zixuan and Shao, Jing},
   booktitle={European conference on computer vision},
   pages={86--103},
  year={2020}
 }

@article{zhang2025dual,
  title={A Dual-Branch CNN for Robust Detection of AI-Generated Facial Forgeries},
  author={Zhang, Xin and Song, Yuqi and Zuo, Fei},
  journal={arXiv preprint arXiv:2510.24640},
  year={2025}
}

@article{farid2009image,
  title={Image forgery detection},
  author={Farid, Hany},
  journal={IEEE Signal processing magazine},
  volume={26},
  number={2},
  pages={16--25},
  year={2009},
  publisher={IEEE}
}

@article{mirsky2021creation,
  title={The creation and detection of deepfakes: A survey},
  author={Mirsky, Yisroel and Lee, Wenke},
  journal={ACM computing surveys (CSUR)},
  volume={54},
  number={1},
  pages={1--41},
  year={2021},
  publisher={ACM New York, NY, USA}
}

@inproceedings{chollet2017xception,
  title={Xception: Deep learning with depthwise separable convolutions},
  author={Chollet, Fran{\c{c}}ois},
  booktitle={Proceedings of the IEEE conference on computer vision and pattern recognition},
  pages={1251--1258},
  year={2017}
}

@inproceedings{tan2019efficientnet,
  title={Efficientnet: Rethinking model scaling for convolutional neural networks},
  author={Tan, Mingxing and Le, Quoc},
  booktitle={International conference on machine learning},
  pages={6105--6114},
  year={2019},
  organization={PMLR}
}

@article{dzanic2020fourier,
  title={Fourier spectrum discrepancies in deep network generated images},
  author={Dzanic, Tarik and Shah, Karan and Witherden, Freddie},
  journal={Advances in neural information processing systems},
  volume={33},
  pages={3022--3032},
  year={2020}
}

@inproceedings{wang2023dire,
  title={Dire for diffusion-generated image detection},
  author={Wang, Zhendong and Bao, Jianmin and Zhou, Wengang and Wang, Weilun and Hu, Hezhen and Chen, Hong and Li, Houqiang},
  booktitle={Proceedings of the IEEE/CVF International Conference on Computer Vision},
  pages={22445--22455},
  year={2023}
}

@inproceedings{chen2024drct,
  title={Drct: Diffusion reconstruction contrastive training towards universal detection of diffusion generated images},
  author={Chen, Baoying and Zeng, Jishen and Yang, Jianquan and Yang, Rui},
  booktitle={Forty-first International Conference on Machine Learning},
  year={2024}
}

@inproceedings{zhang2019detecting,
  title={Detecting and simulating artifacts in GAN fake images},
  author={Zhang, Xu and Karber, Svebor and Chang, Shih-Fu},
  booktitle={WIFS},
  year={2019}
}

@article{schwarz2021frequency,
  title={On the frequency bias of generative models},
  author={Schwarz, Katja and Liao, Yiyi and Geiger, Andreas},
  journal={Advances in Neural Information Processing Systems},
  volume={34},
  pages={18126--18136},
  year={2021}
}

@inproceedings{howard2019searching,
  title={Searching for mobilenetv3},
  author={Howard, Andrew and Sandler, Mark and Chu, Grace and Chen, Liang-Chieh and Chen, Bo and Tan, Mingxing and Wang, Weijun and Zhu, Yukun and Pang, Ruoming and Vasudevan, Vijay and others},
  booktitle={Proceedings of the IEEE/CVF international conference on computer vision},
  pages={1314--1324},
  year={2019}
}

@inproceedings{gilles2025xcepknn,
  title={XcepKNN: Leveraging Hybrid Deep Learning for Enhanced MRI-Based Brain Tumor Classification},
  author={Gilles, Ethan and Song, Yuqi and Zhang, Xin and Zuo, Fei},
  booktitle={2025 IEEE/ACIS 23rd International Conference on Software Engineering Research, Management and Applications (SERA)},
  pages={303--308},
  year={2025},
  organization={IEEE}
}

@inproceedings{xie2017aggregated,
  title={Aggregated residual transformations for deep neural networks},
  author={Xie, Saining and Girshick, Ross and Doll{\'a}r, Piotr and Tu, Zhuowen and He, Kaiming},
  booktitle={Proceedings of the IEEE conference on computer vision and pattern recognition},
  pages={1492--1500},
  year={2017}
}

@incollection{koonce2021efficientnet,
  title={EfficientNet},
  author={Koonce, Brett},
  booktitle={Convolutional neural networks with swift for Tensorflow: image recognition and dataset categorization},
  pages={109--123},
  year={2021},
  publisher={Springer}
}

@article{xinzhang,
  title={RCDN: Real-Centered Detection Network for Robust Face Forgery Identification},
  author={McCurdy, Wyatt and Zhang, Xin and Song, Yuqi and Gao, Min},
  journal={arXiv preprint arXiv:2601.12111},
  year={2026}
}

\end{document}